\pdfoutput=1

\documentclass[11pt]{article}

\usepackage[preprint]{acl}

\usepackage{times}
\usepackage{amsmath}
\usepackage{latexsym}
\usepackage{enumitem}
\usepackage{titlesec}
\usepackage{booktabs}
\usepackage{tabularx}
\usepackage{adjustbox}
\usepackage{float}
\usepackage{caption} 
\titleformat{\paragraph}
{\normalfont\normalsize\bfseries}{\theparagraph}{1em}{}

\titlespacing*{\paragraph}
{0pt}{3.25ex plus 1ex minus .2ex}{1.5ex plus .2ex}

\titleformat{\subparagraph}
  {\normalfont\normalsize\bfseries}{}{0pt}{} 
\titlespacing*{\subparagraph}
  {0pt}{3.25ex plus 1ex minus .2ex}{1.5ex plus .2ex}

\usepackage[T1]{fontenc}

\usepackage[utf8]{inputenc}

\usepackage{microtype}

\usepackage{inconsolata}

\usepackage{graphicx}

\title{Enhancing Marker Scoring Accuracy through Ordinal Confidence Modelling in Educational Assessments\thanks{This is a preprint accepted to ACL 2025 Industry Track. The official publication will appear in the ACL Anthology.}}

\author{
    \textbf{Abhirup Chakravarty},
    \textbf{Mark Brenchley},
    \textbf{Trevor Breakspear},
    \textbf{Ian Lewin},
    \textbf{Yan Huang}
    \\
    \\
    Applied AI
    \\
    Cambridge University Press \& Assessment
    \\
    \small{
        \textbf{Correspondence:} \href{mailto:abhirup.chakravarty@cambridge.org}{abhirup.chakravarty@cambridge.org}, \href{mailto:yan.huang@cambridge.org}{yan.huang@cambridge.org}
    }
  }

\begin{document}
\maketitle
\begin{abstract}
    A key ethical challenge in Automated Essay Scoring (AES) is ensuring that scores are only released when they meet high reliability standards. Confidence modelling addresses this by assigning a reliability estimate measure, in the form of a confidence score, to each automated score. In this study, we frame confidence estimation as a classification task: predicting whether an AES-generated score correctly places a candidate in the appropriate CEFR level. While this is a binary decision, we leverage the inherent granularity of the scoring domain in two ways. First, we reformulate the task as an $n$-ary classification problem using score binning. Second, we introduce a set of novel \textit{Kernel Weighted Ordinal Categorical Cross Entropy (KWOCCE)} loss functions that incorporate the ordinal structure of CEFR labels. Our best-performing model achieves an F1 score of 0.97, and enables the system to release 47\% of scores with 100\% CEFR agreement and 99\% with at least 95\% CEFR agreement---compared to $\approx92\%$ CEFR agreement from the standalone AES model where we release all AM predicted scores.
\end{abstract}

\section{Introduction}
\label{sec:intro}

Automated Essay Scoring (AES) systems aim to evaluate the quality of candidate writing using computational methods. These systems are increasingly adopted in large-scale assessments due to their speed, consistency, and scalability \cite{Xuetal2020, Lottridge_etal2023, ShermisWilson_2024, XuEtAl2024}. A common goal is to assign a proficiency level based on frameworks such as the Common European Framework of Reference (CEFR) \cite{Council_of_Europe_2001}, which defines levels from A1 (beginner) to C2 (advanced). Unlike traditional classification tasks, these levels are ordinal\textemdash with the levels ranked in terms of increasing levels of proficiency.

To enhance accuracy in high-stakes settings, many AES systems adopt a hybrid marking system, where a separate confidence model evaluates the automarker score for a response and only releases a score when it meets a minimum confidence threshold \cite{Xuetal2021, SinglaEtAl2022, VecchioEtAl2018-confidence-for-spoken-assessment}. However, confidence modelling in AES remains underexplored. Most current methods rely on standard regression or classification approaches \cite{johan-berggren-etal-2019-regression}, and while some work has considered the ordinal nature of AES \cite{johan-berggren-etal-2019-regression, mathias-bhattacharyya-2020-neural}, very few have applied ordinal techniques to confidence estimation \cite{Malinin2017, VecchioEtAl2018-confidence-for-spoken-assessment, Loukina2019, funayama-etal-2020-preventing, Gao2024, Orwat2024}.

In this paper, we show how redefining the classification approach and adopting innovative \textit{ordinal loss functions} can optimise confidence model performance. We begin by framing the task as a binary classification problem: predicting whether the AES system score places candidates in the correct CEFR grade. We introduce an increase in granularity, which allows us to explore how fine-grained information impacts confidence estimation and score release decisions, through two extensions: (1) an $N$-ary CEFR classification that estimates the full probability distribution over CEFR levels, and (2) a score-binning approach with $N$-ary classification at the score level, which groups continuous scores into interpretable bins aligned with human marking tolerances. Finally, we introduce a novel loss function\textemdash \textsc{Kernel Weighted Ordinal Categorical Cross-Entropy (KWOCCE)}\textemdash which penalises misclassifications based on the distance between predicted and examiner CEFR levels, building on foundational work by \citet{Frank2001}, and more recent studies that incorporate class distances into loss functions to yield better-calibrated and more robust models \cite{DELATORRE2018144, castagnos-etal-2022-simple, POLAT2025126372}.

KWOCCE generalises prior approaches such as Class Distance Weighted Cross-Entropy \cite{POLAT2025126372} and log-based ordinal losses \cite{castagnos-etal-2022-simple}, enabling exploration of linear, logarithmic, exponential, and Gaussian penalty schemes. The goal is to penalise large misclassifications more heavily while tolerating minor disagreements, aligning with real-world marking practice.

We evaluate our approach in a human-in-the-loop Hybrid Marking System (HMS), where an LLM-based AES engine generates scores and a downstream confidence model determines whether scores are released or escalated for review. To assess real-world utility, we report the percentage of AES scores that can be released at different thresholds of minimum CEFR agreement. Our results show that the proposed KWOCCE loss significantly improves control over score release decisions: up to $\approx47\%$ of AES scores can be released with 100\% CEFR agreement, and up to $\approx99\%$ with at least 95\% CEFR agreement, compared to $\approx92\%$ CEFR agreement from the unaided AES system, where all predicted scores are released.

\vspace{0.5em}
\noindent\textbf{Contributions:}
\begin{itemize} 
    \item We demonstrate the importance of granularity in confidence modelling.
    \item We frame AES confidence estimation as an ordinal classification problem, leveraging the structure of CEFR labels.
    \item We propose the KWOCCE loss, incorporating kernel-based distance penalties into the cross-entropy objective.
    \item We show that KWOCCE improves confidence calibration and score release reliability over standard approaches, supporting safer and more robust AES deployment.
\end{itemize}

This work connects AES to broader advances in ordinal classification and NLP, responding to calls for better alignment between machine predictions and human assessment standards \cite{amigo-etal-2020-effectiveness, castagnos-etal-2022-simple}, integrating methods from uncertainty estimation, ordinal classification, and kernel-based loss design to improve scoring reliability and trustworthiness. 

\section{Background}

Despite growing interest in AES, few studies explicitly address both scoring and confidence estimation. AES is often framed as a standard regression or classification task \cite{johan-berggren-etal-2019-regression, mathias-bhattacharyya-2020-neural}, where confidence is assumed to be reflected by outputs like softmax probabilities or prediction intervals. However, these are not always well-calibrated and may fail to capture real-world reliability\textemdash particularly in high-stakes educational contexts.

One reason for this gap may be the focus on accuracy as the key metric in AI benchmarks, often at the expense of prediction confidence and calibration \cite{Banachewicz2022}. In response to fairness and out-of-domain concerns, some commercial systems prioritise aberrancy detection over intrinsic confidence modelling \cite{Loukina2019, Gao2024}. Earlier solutions combined automated and human marking \cite{Burstein2013}, but this adds cost and sidesteps the core issue of model uncertainty. 

Recent work has explored confidence estimation in deep neural networks, especially when no natural confidence score is available. \citet{Malinin2017} and \citet{VecchioEtAl2018-confidence-for-spoken-assessment} used ensembles and synthetic data to model uncertainty and detect out-of-distribution inputs. \citet{SinglaEtAl2022} showed that confidence modelling can help decide when to escalate AES responses, highlighting that some low-confidence errors are more critical due to their impact on final candidate results.

This issue becomes particularly salient in scenarios where scores are not only assigned but also \textit{banded} into levels depending on which band of scores the AES score lies in, such as the CEFR framework used in second language assessments \cite{Council_of_Europe_2001}. In such settings, errors near band boundaries (e.g., predicting B1 instead of B2) may have a disproportionate effect on outcomes, and thus merit different treatment from errors within a band. Confidence modelling, in this context, must therefore consider not only the likelihood of error but also the potential impact of that error \cite{Orwat2024}.

Beyond the assessment community, the NLP field has begun exploring ordinal classification and distance-aware loss functions as tools for improving confidence calibration. \citet{castagnos-etal-2022-simple} introduced a log-based loss that penalises distant misclassifications more heavily, enhancing both accuracy and interpretability. \citet{POLAT2025126372} proposed a class-distance-weighted cross-entropy for medical severity classification, while \citet{DELATORRE2018144} adapted the weighted Kappa metric into a loss for ordinal deep learning. These works show the benefits of aligning model objectives with ordinal label structure\textemdash especially when near-miss predictions carry partial credit. However, such approaches remain rare in AES.

In this work, we extend the literature by developing a hybrid marking system (HMS) that incorporates kernel-weighted ordinal classification for confidence modelling in AES. Our approach builds on insights from assessment, uncertainty estimation, and NLP tasks of an ordinal nature to propose a principled, loss-driven strategy for score release: only high-confidence predictions\textemdash determined by both prediction certainty and ordinal agreement\textemdash can be released without human review (unless also separately flagged by ancillary aberrant detection systems). This strikes a balance between automation and rigour.

\section{Data}
\label{sec:data}

This study uses a proprietary dataset from a high-stakes second-language English exam. Candidates write two extended responses, each analytically scored by a certified examiner on a 0–20 scale. 
The scores for both parts are summed to produce a component-level score out of 40 and then mapped using proprietary cut scores to one of three possible CEFR levels for the target proficiency band of the exam (CoE, 2001).

Examiner scores and CEFR levels represent a qualitative assessment of learner's second language proficiency relative to the CEFR, providing an overall judgement of writing quality. 

Training and evaluation sets were selected using stratified random sampling to reflect the empirical score distribution and candidate demographics \cite{AERA2014, Lottridge2020, McCaffrey2022, XuEtAl2024}. As a result of the empirical distribution, both raw scores and CEFR levels follow an approximately normal distribution 

The confidence modelling approaches explored in this paper are model agnostic, in that they can be trained and applied to any automarker model. To provide a baseline for assessing the performance of the reported confidence models, we additionally trained a bespoke automarker. This is a transformer-based encoder model with a regression head, trained on 100,000 test-specific responses, with a validation set of 25,000 responses. The confidence model used a disjoint, larger training set of 231,603 responses, with a validation set of 57,901 responses, capturing variance in the automarker while avoiding task overlap. The final evaluation set consists of 644 responses from 322 candidates, in line with prior commercial AES sample sizes \cite{BennetEtAl2015, ShermisEtAl2022, FirooziEtAl2023}. A gold-standard reference score was created via a multi-marking exercise: 15 certified examiners rated all responses, and a fair average (FA) score was derived using Multi-Faceted Rasch Measurement to account for rater effects \cite{Wolfe2004, XuEtAl2024}. 

For evaluation purposes, we report two directly interpretable, domain specific agreement metrics, both computed at the component level (sum of the two part level scores), where candidate outcomes are determined. The first metric, RMSE, is reported on a 0–40 scale and reflects raw score agreement based on the sum of scores across both of the candidates’ two test responses. The second metric, \% CEFR Agreement, is an accuracy-based measure of categorical agreement, capturing the percentage of cases where the automarker assigns the same CEFR level as the FA reference score. This metric focuses on agreement in the final outcome for the examinee, which is critical for high-stakes decision-making. We use CEFR agreement over any other metrics such as QWK, because it has better interpretabilitiy for operational use \cite{di-eugenio-glass-2004-squibs, Pontius10082011, yannakoudakis-cummins-2015-evaluating, Xuetal2021}.

\subsection{Baseline Automarker Performance}
Table \ref{table:raw_performance} shows baseline automarker (AM) performance, assuming 100\% of predicted scores are released (i.e., no confidence model is applied to filter outputs). The AM predicts scores for part-level responses. At the component level (summing up the scores from the two parts), it performs well, achieving an RMSE of 1.09 and CEFR agreement of 91.61\% with the fair-average reference scores. That is, the AM’s predicted scores already closely align with the ground-truth CEFR levels. However, despite the high agreement, there remains room for improvement—particularly in controlling which scores are released, which is critical for high-stakes applications.

\begin{table}[htb]
    \centering
    \resizebox{\linewidth}{!}{
        \begin{tabular}{lcc}
             \toprule
             Comparison Type & RMSE & CEFR Agreement \\
             \midrule
             Raw\footnotemark[1] Automarker & 1.095 & 91.61 \\
             \bottomrule
        \end{tabular}
    }
    \caption{Raw performance of Auto-marker}
    \label{table:raw_performance}
\end{table}
\footnotetext[1]{\textit{Raw} refers to scores assigned without additional QA filtering.}

\section{Experiments}
\label{sec:experiments}
As described in Section \ref{sec:intro}, we frame the problem as a binary classification task: determining whether an automarked score is confident or not in predicting the expected CEFR level for a candidate. Our approach progressively refines the confidence modelling by leveraging the granularity of scoring data.

\subparagraph{Hybrid Marking System (HMS) Framework}
The proposed HMS features an AM and a downstream confidence model. The AM outputs the score for a candidate response and also generates LLM embeddings. The confidence model subsequently uses these embeddings, AM scores, and the CEFR cut scores to predict confidence on a 0–1 scale, with 1 indicating full confidence that the predicted score for a particular response agrees with the expected CEFR level. 

By integrating the AM with a confidence model, the HMS enables nuanced human scoring where confidence is low, helping underpin assessment accuracy and reliability. The confidence model determines whether the generated automarker score is released or the response instead flagged for human review based on a predefined confidence threshold. Designed for diverse assessment contexts, including high-stakes testing and formative evaluation, HMS ensures both precision and adaptability.

\subsection{Experiment 1: Core Architecture}
The confidence model was developed through iterative refinements aimed at improving confidence score assignment for AM predictions. Initial models used simple correctness-based measures, while later versions incorporated statistical insights into model behaviour and score distributions.

The following subsections describe each stage of this progression.

\subsubsection{Binary Classification}
The first approach framed confidence estimation as a binary classification task, labelling each prediction as correct (1) or incorrect (0) based on alignment between the AM score and the true CEFR level. Using Cross-Entropy (CE) loss, the final probability output was interpreted as the confidence score. While simple and interpretable, this baseline lacked granularity in uncertainty estimation.

\subsubsection{CEFR-Level N-ary Classification}
Further analysis showed that AM performance varied across the score range, with greater reliability in data-rich regions. We therefore moved to an $N$-ary classification model, where $N$ is the number of CEFR levels. Using Categorical Cross-Entropy (CCE) loss, the model produced a probability distribution over CEFR levels. Confidence was taken as the probability assigned to the CEFR predicted by the AM. This formulation offered more nuanced uncertainty estimates, particularly in cases with competing CEFR probabilities.

\subsubsection{Score-Level Binned N-ary Classification}
To further increase granularity, we extended the $N$-ary classification by treating individual score points as separate classes. We then applied binning based on CEFR cut scores, summing probabilities of score points within each CEFR band to compute cumulative confidence. The confidence score was derived similarly to the CEFR-level model but benefited from finer resolution, better capturing subtle variations in AM reliability across the score spectrum.

\subsubsection{Core Architecture Results}
\begin{table}[htb]
    \centering
    \resizebox{\linewidth}{!}{
        \begin{tabular}{lcccc}
            \toprule
            Classifier Type & Accuracy & Precision & Recall & F1 \\
            \midrule
            Binary & 0.578 & 0.579 & 0.997 & 0.733 \\
            CEFR $N$-ary & 0.642 & 0.693 & 0.869 & 0.772 \\
            Score Binned $N$-ary & 0.913 & 0.913 & 1.000 & 0.954 \\
            \bottomrule
        \end{tabular}
    }
    \caption{Comparison of classifier performance across architectures}
    \label{table:core_arch_results}
\end{table}

We performed a threshold analysis on the confidence scores generated by each architecture, using a thousand increments. Here, a true-positive would be when a confidence score is above threshold and the predicted score corresponds to the expected CEFR level. A true-negative would be when both the confidence is below threshold and there is a mismatch with respect to the fair average CEFR level. Metrics reported in Table~\ref{table:core_arch_results} correspond to the threshold yielding the best F1 score. Results show consistent improvement with increasing classification granularity, likely due to richer input information and greater tolerance for near-miss predictions. Consequently, the cumulative CEFR probability approach offers a more robust basis for downstream confidence estimation. We adopt the Score Binned $N$-ary classifier as the standard for subsequent experiments.

\subsection{Experiment 2: Ordinal Category Classification (OCC)}
Given the ordinal nature of the problem, we incorporated ordinal relationships into our classification framework. Our OCC benchmark was established using Keras’ OCC loss \cite{JHart96_keras_ordinal_categorical_crossentropy}. Additionally, we developed the Kernel Weighted Ordinal CCE (KWOCCE) loss function to enforce ordinal constraints, better capturing the inherent ordering information.

\subsubsection{Keras OCC Loss}
This loss function extends the standard Categorical Cross-Entropy by introducing a weighting mechanism that penalises predictions based on their distance from the true class. The mathematical formulation of the OCC loss is as follows:
\begin{equation}
\text{loss}(\mathbf{y}, \hat{\mathbf{y}}) = (w(\mathbf{y}, \hat{\mathbf{y}}) + 1) \cdot \operatorname{CE}(\mathbf{y}, \hat{\mathbf{y}})
\end{equation}
\begin{equation}
\quad w(\mathbf{y}, \hat{\mathbf{y}}) = \frac{|\arg\max_i \mathbf{y} - \arg\max_i \hat{\mathbf{y}}|}{K - 1}
\end{equation}

Here, \( K \) represents the total number of classes, \( \mathbf{y} \) is the one-hot encoded true class vector, \( \hat{\mathbf{y}} \) denotes the predicted probability vector and \( \operatorname{CE}(\mathbf{y}, \hat{\mathbf{y}}) \) is the standard cross-entropy loss. The weighting factor \( w \) scales the loss proportionally to the absolute difference between the predicted and true class indices, normalised by \( K - 1 \). This approach ensures that misclassifications closer to the true class incur a lower penalty than those further away, effectively capturing the ordinal nature of the categories.

\subsubsection{KWOCCE}

Keras' OCC loss penalises misclassifications based on distance from the true class using linear scaling, assuming that all ordinal gaps carry equal severity. However, in practice, not all errors are equally consequential; e.g., misclassifying CEFR level 1 as level 2 is less severe than as level 5. To better reflect such distinctions, we propose \textsc{Kernel Weighted Ordinal Categorical Cross-Entropy (KWOCCE)}: a family of loss functions that apply nonlinear, distance-aware penalties via kernel functions. These refinements improve ordinal classification, enhance robustness, and yield more interpretable confidence estimates.

\paragraph{Kernel Functions}

Each kernel function determines how severely a misclassification is penalised based on its distance from the true class. Unlike fixed linear weights, kernel-based schemes allow more nuanced penalisation that aligns with the ordinal structure of CEFR scores. We define \( x = \hat{y} - y \), where \( \hat{y} \) is the predicted class and \( y \) is the true class, and \( N \) is the number of classes, and $\alpha$ and $\beta$, where applicable, are tuned hyperparameters.

\subparagraph{Linear}
    \begin{equation}
        K_{\text{linear}}(x, N) = \max\left(0, 1 - \frac{|x|}{N}\right)
    \end{equation}
    The linear kernel provides a straightforward extension of the Keras OCC loss by scaling penalties proportionally to the absolute classification error. It maintains consistency with ordinal relationships, it does not distinguish between large and small misclassifications beyond the direct ordinal gap.

\subparagraph{Logarithmic}
    \begin{equation}
        K_{\text{log}}(x, N; \alpha) = \max\left(0, 1 - f_{\text{log}}(x, N; \alpha)\right)
    \end{equation}
    \begin{equation}
        f_{\text{log}}(x, N; \alpha) = \frac{\alpha \log(1 + |x|)}{\log(N)}
    \end{equation}
    The logarithmic kernel introduces a progressively decreasing penalisation for larger errors. This function better reflects real-world grading practices, where extreme misclassifications are rare but possible, and minor deviations should not be overly penalised. This approach is particularly useful in settings where small deviations (e.g., 1 to 2) are common and tolerable, whereas larger deviations (e.g., 1 to 5) should still be significantly penalised.

\subparagraph{Exponential}
    \begin{equation}
        K_{\text{exp}}(x; \alpha, \beta) = \max\left(0, f_{\text{exp}}(x; \beta)\right)
    \end{equation}
    \begin{equation}
        f_{\text{exp}}(x; \beta) = \alpha \left(1 -\frac{1}{1 + \exp(\beta - |x|)}\right)
    \end{equation}
        
    The exponential kernel provides a sharper distinction between minor and severe errors. This function assigns minimal penalties to near-correct predictions, while exponentially increasing penalties for larger misclassifications. This is particularly useful in high-stakes assessment settings, where confidence in high-accuracy predictions is crucial.

\subparagraph{Gaussian}
    \begin{equation}
        K_{\text{gaussian}}(x; \alpha) = \max\left(0, f_{\text{exp}}(x; \alpha)\right)
    \end{equation}
    \begin{equation}
        f_{\text{exp}}(x; \alpha) = \exp\left(-\left(\frac{x}{\alpha}\right)^2\right)
    \end{equation}

    The Gaussian kernel applies a bell-shaped penalty, ensuring that small classification errors are barely penalised, while large errors receive exponentially higher penalties. This model best aligns with human grading behaviour, where minor misjudgements are tolerated, but gross errors significantly impact the assigned CEFR.

\paragraph{Kernel-Weighted Cross-Entropy Loss}

To integrate the kernel weighting into our classification framework, we modify the standard cross-entropy loss function to account for ordinal misclassification penalties. This ensures that correct or near-correct predictions incur lower penalties, while distant misclassifications are progressively penalised according to the chosen kernel.

\begin{equation}  
\mathcal{L}(\mathbf{y}, \hat{\mathbf{y}}) = -\sum_{i=1}^N w_i \log \hat{y}_{i c_i}
\end{equation}  

Here, \( \mathbf{y} \) is the true one-hot label, \( \hat{\mathbf{y}} \) is the predicted probability vector, \( c_i \) is the true class index, and \( w_i \) is the kernel-derived penalty based on the distance between predicted and true classes.

\paragraph{Reduction Method}

The final loss value is calculated using a mean reduction approach. This computes the average loss across all samples, ensuring that the gradients remain stable and are not dominated by a small subset of extreme misclassifications. 

\begin{equation}
\mathcal{L}_{\text{mean}} = \frac{1}{N} \sum_{i=1}^N \mathcal{L}_i
\end{equation}

\begin{table*} [htbp]
    \centering
    \renewcommand{\thetable}{4} 
    \captionsetup{labelformat=default,labelsep=colon} 
    \resizebox{\linewidth}{!}{%
        \begin{tabular}{lrrrrrr}
            \toprule
            Model & Precision & Recall & F1-Score & F0.5-Score & Accuracy & AUC-ROC \\
            \midrule
            Benchmark & 0.913 & 1.000 & 0.954 & 0.929 & 0.913 & 0.848 \\
            Keras OCC & 0.935 & 1.000 & 0.966 & 0.947 & 0.935 & 0.793 \\
            KWOCCE Linear & 0.935 & 1.000 & 0.966 & 0.947 & 0.935 & 0.557 \\
            KWOCCE Log$_{\alpha = 3}$ & 0.936 & 1.000 & 0.967 & 0.948 & 0.936 & 0.755 \\
            KWOCCE Exp$_{(\alpha = 1, \beta = 3)}$ & 0.938 & 0.998 & 0.967 & 0.949 & 0.936 & 0.738 \\
            KWOCCE Gaussian$_{\alpha = 0.5}$ & 0.936 & 1.000 & 0.967 & 0.948 & 0.936 & 0.806 \\
            \bottomrule
        \end{tabular}
    }
    \caption{Model Binary Classification Metrics}
    \label{table:nlp_stats_combined}
\end{table*}

\subsubsection{OCC Results}
\begin{table}[H]
    \centering
    \renewcommand{\thetable}{3} 
    \captionsetup{labelformat=default,labelsep=colon} 
    \resizebox{\linewidth}{!}{
        \setlength{\tabcolsep}{5pt}
        \begin{tabular}{lrrrr}
            \toprule
            Loss function & \multicolumn{2}{c}{100\% CEFR Agree} & \multicolumn{2}{c}{95\% CEFR Agree} \\
            \cmidrule(lr){2-3} \cmidrule(lr){4-5}
            & RMSE & \% Release & RMSE & \% Release \\
            \midrule
            Benchmark & 0.912 & 29.80 & 1.143 & 91.83 \\
            Keras OCC & 0.854 & 36.31 & 1.049 & 91.97 \\
            KWOCCE Linear & 1.006 & 47.35 & 1.068 & 98.16 \\
            KWOCCE Log$_{\alpha = 3}$ & 0.854 & 19.86 & 1.057 & 98.89 \\
            KWOCCE Exp$_{(\alpha = 1, \beta = 3)}$ & 0.964 & 41.01 & 1.062 & 99.12 \\
            KWOCCE Gaussian$_{\alpha = 0.5}$ & 0.940 & 35.73 & 1.057 & 98.75 \\
            \bottomrule
        \end{tabular}
    }
    \caption{Comparison of OCC Loss Performance at 100\% and 95\% CEFR Agreement Thresholds}
    \label{table:combined_loss_performance}
\end{table}

All OCC models were evaluated using standard NLP metrics as well as domain-specific validation metrics to better assess real-world impact. Our primary validation metric is the percentage of AM scores that can be released under each model for a particular threshold of CEFR agreement. We operationalise this as the percentage of exact CEFR agreement achieved with our gold-standard fair average (FA) reference. 
More specifically, at each confidence threshold, we identify the particular set of automarker scores that are “high confidence” (i.e. those that are at or above the confidence threshold). These high confidence automarker scores are then swapped in over the corresponding FA scores and used to determine a revised set of CEFR levels. Finally, the resulting level of agreement is calculated by comparing the overlap between this revised set of CEFR levels and the CEFR level achieved if no automarker scores had been released and candidates received only Fair Average scores.

Table~\ref{table:combined_loss_performance} compares the performance of different confidence models at two thresholds: a maximum of 100\% agreement and a minimum of 95\%. Both represent meaningful improvements over the AM’s unaided agreement level of $\approx92\%$.

At 100\% CEFR agreement, the best RMSE values are achieved by Keras OCC (0.8544) for 36.31\% released and KWOCCE Log ($\alpha = 3$) (0.8537) for 19.86\% released, indicating that these methods produce the most reliable confidence scores. RMSE remains relatively stable across models, and always lower than the unaided AM RMSE (1.095), suggesting that the confidence mechanism helps reduce grading variance when the system is more certain.

KWOCCE Linear achieves the highest percentage of AM scores released (47.35\%), indicating its ability to more confidently identify and correctly classify high-certainty responses. This suggests stronger alignment between the model’s confidence scores and the ground-truth CEFR labels.

At 95\% CEFR agreement, all KWOCCE variants outperform both Keras and Benchmark baselines in every metric except RMSE. However, in this setting, RMSE is considered a secondary metric\textemdash our primary concern is accurate CEFR assignment. Small RMSE variations are tolerable as long as they remain substantively low and better than the unaided AM RMSE. Performance for intermediate thresholds between 99\% and 96\% CEFR agreement is reported in Appendix~\ref{sec:appendix::cefr_perf}.

Table~\ref{table:nlp_stats_combined} presents results for the final downstream binary classification task: determining whether the model is confident in the CEFR agreement of AM scores. While the benchmark model using standard CCE loss achieves high AUC-ROC and perfect recall, these metrics alone are insufficient. Precision, F1, F0.5, and Accuracy suggest that explicitly modelling ordinal structure leads to better convergence and more reliable decision-making. Performance on the original CEFR-level classification task can be found in Appendix~\ref{sec:appendix::nlp_stats}.

\section{Conclusion}
Our experiments show that the most granular architecture\textemdash the Score-level Binned $N$-ary Classifier\textemdash consistently performs best. A clear trend emerges: increasing granularity improves confidence modelling. These gains are evident across standard NLP metrics (Precision, Recall, F1, F0.5, AUC-ROC, and Accuracy) and domain-specific validation metrics, such as the \% AM released at different CEFR agreement thresholds. 

Our findings show that a candidate’s likelihood of receiving the appropriate outcome is best determined by models that respect the domain’s ordinal structure—leveraging raw score information, the inherent order of CEFR labels, and KWOCCE loss functions that penalise large misclassifications more heavily. Our best-performing model (\textit{KWOCCE Linear}) enabled the release of up to $\approx47\%$ of scores with 100\% CEFR agreement, and up to $\approx99\%$ with at least 95\% CEFR agreement—compared to $\approx92\%$ CEFR agreement from the unaided AM system, which released 100\% of scores with no confidence control. Thus, we achieve our goal of greater control over score release, leading to higher operational reliability, while still enabling greater volumes of automarker scores to be released in principle—resulting in a more favourable trade-off between coverage and reliability. The refined control enabled by fine-grained confidence modelling offers a promising step towards more ethical and effective automated test scoring.

\section*{Limitations}
The model used in this preliminary study was trained and evaluated on data from a single exam with a particular proficiency distribution. Although the evaluation dataset is multi-marked, representative, and comparable in size to other commercial AES datasets, it remains relatively small compared to test sets in other domains. Future work will assess the efficacy of the novel functions on models trained using a wider range of simulated and operational data, as well as evaluated using larger datasets as well as including data from other exams. 

\bibliography{anthology,references}

\appendix
\label{sec:appendix}

\section{CEFR Agreement threshold metrics}
\label{sec:appendix::cefr_perf}
\begin{table}[H]
    \centering
    \resizebox{\linewidth}{!}{
        \setlength{\tabcolsep}{3pt}
        \begin{tabular}{lrr}
            \toprule
            Loss function & RMSE & \% Released \\
            \midrule
            Benchmark & 1.022 & 61.65 \\
            Keras OCC & 1.025 & 69.46 \\
            KWOCCE Linear & 1.046 & 68.24 \\
            KWOCCE Log$_{\alpha = 3}$ & 1.031 & 65.10 \\
            KWOCCE Exp$_{(\alpha = 1, \beta = 3)}$ & 1.025 & 69.36 \\
            KWOCCE Gaussian$_{\alpha = 0.5}$ & 1.034 & 64.35 \\
            \bottomrule
        \end{tabular}
    }
    \caption{Loss Performance at 99\% CEFR Agreement}
    \label{table:loss_99_cefr}
\end{table}

\begin{table}[H]
    \centering
    \resizebox{\linewidth}{!}{
        \setlength{\tabcolsep}{3pt}
        \begin{tabular}{lrr}
            \toprule
            Loss function & RMSE & \% Released \\
            \midrule
            Benchmark & 1.031 & 66.04 \\
            Keras OCC & 1.028 & 79.28 \\
            KWOCCE Linear & 1.020 & 74.27 \\
            KWOCCE Log$_{\alpha = 3}$ & 1.022 & 74.55 \\
            KWOCCE Exp$_{(\alpha = 1, \beta = 3)}$ & 1.035 & 75.03 \\
            KWOCCE Gaussian$_{\alpha = 0.5}$ & 1.021 & 74.61 \\
            \bottomrule
        \end{tabular}
    }
    \caption{Loss Performance at 98\% CEFR Agreement}
    \label{table:loss_98_cefr}
\end{table}
\begin{table}[H]
    \centering
    \resizebox{\linewidth}{!}{
        \setlength{\tabcolsep}{3pt}
        \begin{tabular}{lrr}
            \toprule
            Loss function & RMSE & \% Released \\
            \midrule
            Benchmark & 1.099 & 74.23 \\
            Keras OCC & 1.016 & 83.58 \\
            KWOCCE Linear & 1.021 & 79.91 \\
            KWOCCE Log$_{\alpha = 3}$ & 1.011 & 81.31 \\
            KWOCCE Exp$_{(\alpha = 1, \beta = 3)}$ & 1.021 & 77.67 \\
            KWOCCE Gaussian$_{\alpha = 0.5}$ & 1.031 & 77.96 \\
            \bottomrule
        \end{tabular}
    }
    \caption{Loss Performance at 97\% CEFR Agreement}
    \label{table:loss_97_cefr}
\end{table}
\begin{table}[H]
    \centering
    \resizebox{\linewidth}{!}{
        \setlength{\tabcolsep}{3pt}
        \begin{tabular}{lrr}
            \toprule
            Loss function & RMSE & \% Released \\
            \midrule
            Benchmark & 1.105 & 83.40 \\
            Keras OCC & 1.039 & 90.59 \\
            KWOCCE Linear & 1.051 & 96.51 \\
            KWOCCE Log$_{\alpha = 3}$ & 1.034 & 83.95 \\
            KWOCCE Exp$_{(\alpha = 1, \beta = 3)}$ & 1.030 & 87.02 \\
            KWOCCE Gaussian$_{\alpha = 0.5}$ & 1.022 & 87.20 \\
            \bottomrule
        \end{tabular}
    }
    \caption{Loss Performance at 96\% CEFR Agreement}
    \label{table:loss_96_cefr}
\end{table}

In Tables \ref{table:loss_99_cefr}, \ref{table:loss_98_cefr}, and \ref{table:loss_97_cefr}, we see that Keras OCC performs reliably well, between 99\% and 97\% CEFR agreements, followed by models trained using KWOCCE losses. In Table \ref{table:loss_96_cefr}, we see that KWOCCE linear outperforms all models by a gap of almost 6\% in the \% AM-released metric. We also see that the OCC functions maintain a stabler lower RMSE than the benchmark, which goes towards the argument of better reliability.

\section{NLP Metrics}
\label{sec:appendix::nlp_stats}

In Table \ref{table:nlp_micro}, the F1 scores (0.9071 for all OCC models) indicate strong correctness when averaged over all classifications. The OCC model scores are consistently higher than the standard benchmark model with CCE loss.

\begin{table} [H]
    \centering
    \resizebox{\linewidth}{!}{
        \setlength{\tabcolsep}{3pt}
        \begin{tabular}{lrrrr}
            \toprule
            Loss function & Precision & Recall & F-1 & F-0.5 \\
            \midrule
            Benchmark & 0.9057 & 0.9057 & 0.9057 & 0.9057 \\
            Keras OCC & 0.9071 & 0.9071 & 0.9071 & 0.9071 \\
            KWOCCE Linear & 0.9071 & 0.9071 & 0.9071 & 0.9071 \\
            KWOCCE Log$_{\alpha = 3}$ & 0.9071 & 0.9071 & 0.9071 & 0.9071 \\
            KWOCCE Exp$_{(\alpha = 1, \beta = 3)}$ & 0.9071 & 0.9071 & 0.9071 & 0.9071 \\
            KWOCCE Gaussian$_{\alpha = 0.5}$ & 0.9071 & 0.9071 & 0.9071 & 0.9071 \\
            \bottomrule
        \end{tabular}
    }
    \caption{Loss Performance: NLP Metrics (Micro)}
    \label{table:nlp_micro}
\end{table}

In Table \ref{table:nlp_macro}, the benchmark model (0.7538 Macro F1) performs best, indicating balanced performance across all class distributions. KWOCCE Linear and KWOCCE Log degrade significantly ($\approx$ 0.57-0.59 Macro F1), suggesting that these methods struggle with minority classes. Keras OCC maintains moderate performance (0.6209 Macro F1), demonstrating a reasonable trade-off.

\begin{table} [H]
    \centering
    \resizebox{\linewidth}{!}{
        \setlength{\tabcolsep}{3pt}
        \begin{tabular}{lrrrr}
            \toprule
            Loss function & Precision & Recall & F-1 & F-0.5 \\
            \midrule
            Benchmark & 0.7538 & 0.6568 & 0.6897 & 0.7226 \\
            Keras OCC & 0.6062 & 0.6486 & 0.6209 & 0.6109 \\
            KWOCCE Linear & 0.5785 & 0.5324 & 0.5386 & 0.5548 \\
            KWOCCE Log$_{\alpha = 3}$ & 0.5706 & 0.6186 & 0.5807 & 0.5726 \\
            KWOCCE Exp$_{(\alpha = 1, \beta = 3)}$ & 0.5951 & 0.6356 & 0.6085 & 0.5993 \\
            KWOCCE Gaussian$_{\alpha = 0.5}$ & 0.5786 & 0.5621 & 0.5685 & 0.5741 \\
            \bottomrule
        \end{tabular}
    }
    \caption{Loss Performance: NLP Metrics (Macro)}
    \label{table:nlp_macro}
\end{table}

In Table \ref{table:nlp_weighted}, the benchmark model retains high precision (0.89), ensuring stable overall classification. KWOCCE Log and Gaussian models maintain moderate generalisation, balancing performance across different CEFR distributions. Keras OCC performs better than KWOCCE and worse than benchmark, keeping the trend consistent, as seen in Table \ref{table:nlp_macro}.

\begin{table} [H]
    \centering
    \resizebox{\linewidth}{!}{
        \setlength{\tabcolsep}{3pt}
        \begin{tabular}{lrrrr}
            \toprule
            Loss function & Precision & Recall & F-1 & F-0.5 \\
            \midrule
            Benchmark & 0.8919 & 0.9010 & 0.8956 & 0.8932 \\
            Keras OCC & 0.8749 & 0.8460 & 0.8588 & 0.8681 \\
            KWOCCE Linear & 0.8507 & 0.8887 & 0.8659 & 0.8552 \\
            KWOCCE Log$_{\alpha = 3}$ & 0.8550 & 0.8604 & 0.8576 & 0.8560 \\
            KWOCCE Exp$_{(\alpha = 1, \beta = 3)}$ & 0.8669 & 0.8539 & 0.8601 & 0.8641 \\
            KWOCCE Gaussian$_{\alpha = 0.5}$ & 0.8553 & 0.8697 & 0.8621 & 0.8579 \\
            \bottomrule
        \end{tabular}
    }
    \caption{Loss Performance: NLP Metrics (Weighted)}
    \label{table:nlp_weighted}
\end{table}

\end{document}